# Higher dimensional homodyne filtering for suppression of incidental phase artifacts in multichannel MRI

Joseph Suresh Paul*, Uma Krishna Swamy Pillai

Medical Image Computing and Signal Processing Laboratory

Indian Institute of Information Technology and Management-Kerala

Trivandrum India

Mailing Address

Dr. Joseph Suresh Paul

Medical Image Computing and Signal Processing Laboratory

Indian Institute of Information Technology and Management-Kerala

Trivandrum India

*j.paul@iiitmk.ac.in

## Abstract

The aim of this paper is to introduce procedural steps for extension of the 1D homodyne phase correction for k-space truncation in all gradient encoding directions. Compared to the existing method applied to 2D partial k-space, signal losses introduced by the phase correction filter is observed to be minimal for the extended approach. In addition, the modified form of phase correction mitigates Incidental Phase Artifacts (IPA) due to truncation. For parallel imaging with undersampling along phase encode direction, the extended homodyne filtering is shown to be effective for minimizing these artifacts when each of the channel k-spaces are truncated along both phase and frequency encode directions. This is illustrated with 2D partial k-space for flow compensated multichannel Susceptibility Weighted Imaging (SWI). Extension of our method to 3D partial k-space shows improved reconstruction of flow information in phase contrast angiography.

**Key words :** 2D partial k-space, 3D partial k-space, phase errors, Incidental Phase Artifacts, ringing artifacts, Parallel imaging

## 1. Introduction

Scan time reduction is achieved by limiting the acquired points in either phase-encode, frequency-encode or both directions, resulting in a partial k-space. A 2D partial k-space is obtained by constraining the signal acquisition in both phase and frequency-encode directions. Under the assumption that the image is real valued, a major part of the unacquired k-space can be filled by imposing the conjugate symmetry constraint [1]. The remaining parts are filled with zeros (truncated). The zero filling leads to truncation artifacts in the resultant image. Broadly, the truncation effects appear in the form of blurring and *Gibbs Ringing* artifacts. The latter is manifested as incidental phase artifacts of sharper image edges, with repeated bands appearing parallel to the edge [2]. The period of repetitions is small in low-resolution images. Hence the problem of *Gibbs Ringing* becomes insignificant in high-resolution images.

In the presence of spatial phase variations, the assumption of a real image is no longer valid. This can arise due to off-resonance effects [3-5], non-centering of echoes in the readout direction [6], or tissue susceptibility variations [7]. In contrast with the real-valued case, a constrained reconstruction is then only possible by replacing all the unacquired k-space points with zeros. This introduces additional errors in the reconstruction, often referred to as *phase errors*. These include an additional component of blur, signal losses, and truncation artifacts. The artifacts appear in the form of localized high frequency intensity variations with repetition intervals smaller than those in *Gibbs ringing.* The artifacts occur in regions with abrupt phase variations [8]. This is demonstrated using a clinical image and validated using numerical simulations.

Phase correction methods such as Homodyne, and Projection On to Convex Sets (POCS) [9-10] have been applied for reduction of phase errors. None of the phase correction methods mentioned above have been successful in restoring the signal loss due to phase errors. Also, application of homodyne filtering is always accompanied by an inherent signal loss due to the filtering operation. Though it causes signal loss and blurring, the ability to remove IPA compared to POCS method, an improved version of homodyne filter accompanied by minimal signal loss will be a desirable solution. In lieu of this idea, we present an extended homodyne phase correction in which the partial k-space is separately weighted along each gradient direction, followed by compensation using phase obtained from low resolution symmetric portion of the k-space along respective gradient direction.

## 2. Background of the problem

To demonstrate the salient features of proposed filtering method, two in vivo examples are used, viz. flow compensated Susceptibility Weighted Image (SWI) and low flip angle gradient echo MRI.  All images are acquired on 1.5T clinical MR scanner (Magnetom- Avanto, Siemens, Erlangen, Germany) with a 12-channel head coil. The Susceptibility Weighted Image is acquired at TE = 23ms, TR = 260ms with 0.25 mm3 in-plane and 5mm3 out-of-plane resolution and the low flip angle GRadient Echo Image (GREI) at TE = 16.4ms, TR = 150ms with 0.25mm3 in-plane and 4mm3 out-of-plane resolution. The phase correction approach is applied on the acquired k-space after offline truncation in both phase and frequency encode directions. The images obtained using Fourier reconstruction applied to the partial k-spaces of SWI and GREI are illustrated in Fig.1(A) and Fig.1(B) respectively. The top part of each panel depicts a graphical representation of the k-space from which the images are reconstructed. Left panels (a) correspond to the full k-space and right panels (b) correspond to the partial k-space. The bottom part of each panel shows the respective magnitude and phase images. For a 1024 × 1024 resolution, the partial k-space is obtained by choosing a 30% fraction of the full k-space, consisting of *n*=40 fractional lines in the phase-encode direction, and *m*=80 frequency points in the frequency-encode direction. For the SWI example of Fig. 1(A), effects of truncation and phase errors is seen in the magnitude image, in the form of blurring and regions with signal losses. For sake of comparison, we have chosen Regions-Of-Interest (ROIs) around the frontal and occipital region. The insets of ROIs highlight areas with signal losses in the magnitude image. For the GREI example of Fig. 1(B), IPA are also observed in the magnitude image obtained from partial k-space reconstruction ( panel (b) ) in addition to the blurring and signal losses. The insets show ROIs highlighting regions affected by IPA.

Fig. 1(A)

Fig. 1(B)

## 3. <u>Numerical Simulation</u>

The IPA artifacts originate from the high-frequency phase components. This is demonstrated using numerical phantoms shown in Fig. 2 and consisting of (a) magnitude image, (b) magnitude image combined with a low frequency phase image, (c) magnitude image in combination with low and high frequency phase components.

Fig. 2

Panels (a)-(c) of Fig. 2 represent the magnitude image, low frequency phase image, and combined low and high frequency phase image respectively. The magnitude image in Fig. 2(a) is constructed using a circular object with uniform intensity. The spatial low frequency phase component is simulated using

$$\phi_{low} = \pi\left(\frac{2r^2}{R_0^2} - 1\right), \qquad 0 < r \le R_0 \tag{1}$$

The high frequency phase image is simulated using a set of concentric discs with uniform phase within each segment, and abrupt transition of phase across the segments. A k-space is constructed for each case (a)-(c). 2-D partial k-spaces for each case are then constructed by truncating in both phase and frequency encoding directions. The unfilled portions are replaced with zeros. The magnitude images obtained by partial Fourier reconstruction is shown in Fig. 3(a-c).

Fig. 3

It is seen that IPA are observed only for the case where the high frequency components are present. The remainder of the two cases exhibits only the *Gibbs ringing* artifacts.

# 4. Proposed Method

## 4.1 Phase correction for 2D partial k-space

For a given 2D-partial k-space $K_{pk}(k_x,k_y)$ with $n$ fractional lines in the phase-encode direction and $m$ frequency points in the frequency-encode direction, the steps used in the extended homodyne phase correction are summarized below.

Step 1 : $K_{pk}(k_x,k_y)$ is independently weighted using $W(k_y)$ and $W(k_x)$ given by

$$WK_{pk1}(k_x,k_y) = K_{pk}(k_x,k_y)\cdot W(k_y) \qquad where \quad W(k_y) = \begin{cases} \frac{k_y}{n}+1 & for\ -n \le k_y \le n, \\ 0 & otherwise. \end{cases} \tag{2}$$

$$WK_{pk2}(k_x,k_y) = K_{pk}(k_x,k_y)\cdot W(k_x) \qquad where \quad W(k_x) = \begin{cases} 1-\frac{k_x}{m} & for\ -m \le k_x \le m, \\ 0 & otherwise. \end{cases} \tag{3}$$

These weighted k-spaces are Fourier transformed to produce the images $s_{pk}(x,y) * w(y)$ and $s_{pk}(x,y) * w(x)$.

Step 2: The phase correction is performed by extracting phase of the low frequency component in the frequency and phase-encode directions independently. This is obtained using

$$K_{sym1}(k_x,k_y) = K_{pk}(k_x,k_y)\cdot \Pi\left(\frac{k_y}{2n}\right), \qquad where \quad \Pi\left(\frac{k_y}{2n}\right) = \begin{cases} 1 & for\ -n \le k_y \le n \\ 0 & otherwise \end{cases} \tag{4}$$

$$K_{sym2}(k_x,k_y) = K_{pk}(k_x,k_y)\cdot \Pi\left(\frac{k_x}{2m}\right), \qquad where \quad \Pi\left(\frac{k_x}{2m}\right) = \begin{cases} 1 & for\ -m \le k_x \le m \\ 0 & otherwise \end{cases} \tag{5}$$

$s_{sym1}$ and $s_{sym2}$ obtained by application of Fourier reconstruction on $K_{sym1}$ and $K_{sym2}$.

Step 3: The complex exponential of these low frequency phase components are then individually multiplied with the respective complex images obtained by application of Fourier reconstruction on pre-weighted partial k-spaces computed in step-1. The corresponding mathematical operations are

$$p_1^*(x,y) = \exp(-i\,angle(s_{sym1}(x,y))) \quad (6)$$

$$p_2^*(x,y) = \exp(-i\,angle(s_{sym2}(x,y))) \quad (7)$$

The phase corrected images are obtained by,

$$s_{comp1} = p_1^*(x,y) \cdot (s_{pk}(x,y) * w(y)) \quad (8)$$

$$s_{comp2} = p_2^*(x,y) \cdot (s_{pk}(x,y) * w(x)) \quad (9)$$

Step 4: The real parts of the resulting complex images $s_{comp1}$ and $s_{comp2}$ are averaged to get final image $s_{comp}$. The steps are summarized in the block schematic shown in Fig. 4.

Fig. 4

## 4.2 Phase correction for 3D partial k-space

A 3D partial k-space is obtained by truncating k-space along all direction. The resultant partial k-space $K_{pk}(k_x,k_y,k_z)$ will have $n$ fractional lines in the phase-encode direction, $m$ frequency points in the frequency-encode direction and $o$ points in slice direction. The steps used in 3D homodyne phase correction are summarized below.

Step 1 : $K_{pk}(k_x,k_y,k_z)$ is independently weighted using $W(k_y)$, $W(k_x)$ and $W(k_z)$ given by

$$WK_{pk1}(k_x,k_y,k_z) = K_{pk}(k_x,k_y,k_z)\cdot W(k_y) \qquad where \quad W(k_y) = \begin{cases} \frac{k_y}{n}+1 & for\ -n \le k_y \le n, \\ 0 & otherwise. \end{cases} \tag{10}$$

$$WK_{pk2}(k_x,k_y,k_z) = K_{pk}(k_x,k_y,k_z)\cdot W(k_x) \qquad where \quad W(k_x) = \begin{cases} 1-\frac{k_x}{m} & for\ -m \le k_x \le m, \\ 0 & otherwise. \end{cases} \tag{11}$$

$$WK_{pk3}(k_x,k_y,k_z) = K_{pk}(k_x,k_y,k_z)\cdot W(k_o) \qquad where \quad W(k_o) = \begin{cases} 1-\frac{k_o}{o} & for\ -o \le k_z \le o, \\ 0 & otherwise. \end{cases} \tag{12}$$

These weighted k-spaces are Fourier transformed and added together to produce the image $s_{wpk}\ (x,y,z)$

Step 2: The phase correction is performed by extracting phase of the low frequency component in the frequency and phase-encode directions independently. This is obtained using

$$K_{sym1}(k_x,k_y,k_z) = K_{pk}(k_x,k_y,k_z)\cdot \Pi\left(\frac{k_y}{2n}\right), \qquad where \quad \Pi\left(\frac{k_y}{2n}\right) = \begin{cases} 1 & for\ -n \le k_y \le n \\ 0 & otherwise \end{cases} \tag{13}$$

$$K_{sym2}(k_x,k_y,k_z) = K_{pk}(k_x,k_y,k_z)\cdot \Pi\left(\frac{k_x}{2m}\right), \qquad where \quad \Pi\left(\frac{k_x}{2m}\right) = \begin{cases} 1 & for\ -m \le k_x \le m \\ 0 & otherwise \end{cases} \tag{14}$$

$$K_{sym3}(k_x,k_y,k_z) = K_{pk}(k_x,k_y,k_z)\cdot \Pi\left(\frac{k_z}{2o}\right), \qquad where \quad \Pi\left(\frac{k_z}{2o}\right) = \begin{cases} 1 & for\ -o \le k_z \le o \\ 0 & otherwise \end{cases} \tag{15}$$

These three symmetric k-spaces added together and fourier transformed to obtain the image $s_{sym}$.

Step 3: The complex exponential of these low frequency phase components are then multiplied with the complex image obtained by application of Fourier reconstruction on pre-weighted partial k-spaces computed in step-1. The corresponding mathematical operations are

$$p^*(x,y,z) = \exp(-i\,angle(s_{sym}(x,y,z))) \qquad (16)$$

The phase corrected image is obtained by,

$$s_{comp} = p^*(x,y,z)\cdot(s_{wpk}) \qquad (17)$$

## 5. <u>Results</u>

### 5.1 Application of homodyne reconstruction filter to 2D partial k-space

Fig.5 illustrates the phase corrected images of low flip angle GREI using POCS, homodyne, and extended homodyne methods. Bottom panels show zoomed version of regions enclosed by rectangular bounding boxes. The leftmost panel shows the zero-filled image without phase correction. The bounding boxes highlight regions with IPA and signal loss due to truncation. The latter regions are enclosed within yellow contours. The result of application of POCS method is shown in panel (b). As discussed in section 2, the IPA are not removed using this method. It is also seen that POCS does not compensate for signal loss due to truncation. Compensated images obtained using homodyne and extended homodyne methods are shown in panels (c)-(d). While the homodyne method is able to remove IPA, it introduces an additional component of signal loss in the region enclosed within the larger yellow contour. The extended homodyne method is able to compensate for this additional signal loss, together with the IPA. Hence the proposed method is a more suitable choice for phase compensation in the presence of IPA.

Fig. 5

Severe signal losses are observed for the flow compensated SWI image shown in Fig.6. For the zero filled reconstruction in panel (a), the effect of truncation appears in the form of blurring and signal losses. These are highlighted in (a1)-(a3) using ROIs chosen from the frontal, brain stem and occipital regions. Panel (b) shows the POCS reconstructed image. Application of conventional homodyne filtering leads to significant signal losses in these areas as shown in panels (c1)-(c3). The results of extended homodyne filtering are shown in panels (d1)-(d3).

Fig. 6

### 5.2 Application of homodyne reconstruction filter to 3D partial k-spaces

Salient features of improved 3D homodyne reconstruction are illustrated using Phase Contrast Magnetic Resonance Angiography (PCMRA) datasets of two volunteers. Both datasets are acquired on 1.5T clinical MR scanner (Magnetom- Avanto, Siemens, Erlangen, Germany) with a 6-channel head coil acquired at TE = 9ms, TR = 56.70 with 10 cm/s velocity encoding. Each dataset consists of 4 partitions with flow encoding first moments altered in pairs. Balanced four point method is used to reconstruct the magnitude flow images along *x*,*y* and *z* directions [11]. The 3D partial k-space for each partition is obtained by offline truncation of the respective k-space. For 167 x 512 x 64 array, 3D partial k-space is synthesized by zero filling each k-space partition as indicated in Fig. 7.

Fig. 7

Each 3D partial k-space is phase corrected using modified homodyne phase correction algorithm described in section 4.2. Balanced four point method is now applied to these phase corrected images. The PD speed is obtained by maximum intensity projection of the combined *x*, *y* and *z* flow images. This is illustrated in Figs. 8A and 8B for the two datasets. Panels (a1)-(c1) show the images reconstructed using zero-filled 3D partial k-space, 3D POCS reconstruction, and the modified 3D homodyne filter. The colored rectangular boxes show regions of interest with improved reconstruction using the proposed method.

Fig. 8A

Fig. 8B

### 5.3 Application to parallel imaging

In parallel imaging, the individual channel k-spaces are generally undersampled along phase encode direction. For uniform undersampling, this takes the form of interleaved k-space lines. Image reconstruction from individual channels is accompanied by aliasing artifacts, dependent on the ratio of undersampling. These artifacts are eliminated using standard reconstruction procedures such as SENSE or GRAPPA [12]. In this section, we present the effect of such reconstruction methods, when the individual channel k-spaces are truncated in the frequency encode direction in addition to interleaving along phase encode direction. Image reconstructed from truncated k-space with an undersampling ratio 4 is shown in Fig. 9. The truncation is performed for a k-space fraction of 0.4. This means that disregarding the undersampling along phase encode direction, only 40% of the total k-space is considered as acquired. The remaining portions are filled with zeros. Further, the k-space lines in the acquired part are intermittently made zeros with a ratio of 1:4 (i.e., for every line retained, 4 subsequent lines are made zeros). Figs. (A) and (B) each correspond to the full images and regions within the areas of interest respectively. The first four column-wise panels represent images from the individual coils. The fifth column shows the combined image. Row-wise panels (a)-(e) show images reconstructed using zero-filled k-space, GRAPPA, GRAPPA followed by POCS, homodyne and extended homodyne respectively. It is seen that GRAPPA compensates only aliasing effects and does not eliminate IPA. The yellow arrows represent the areas with improved reconstruction and reduced signal losses.

Fig. 9A

Fig. 9B

## 6. Discussion and Summary

This work presents an extended homodyne phase correction method which works well for images with IPA. The extended filter mitigates artifacts introduced by k-space truncation due to incidental phase variations. Existing approaches for homodyne phase correction is also effective in removing these artifacts, but introduces signal losses in addition to those resulting from truncation. It is shown that the extended homodyne filter can compensate for these losses. In the absence of IPA, phase correction using extended homodyne method is compatible with that of POCS.

The effect of incidental phase artifacts are further illustrated using introduction of high frequency components into the phase image. This is achieved by first extracting the high frequency phase of the numerically simulated example in section 3. Varying degrees of incidental phase are then numerically simulated using the model

$$\phi_{\lambda}(x,y) = (1-\gamma)\phi_{low}(x,y) + \gamma\phi_{high}(x,y), \qquad 0 \le \gamma \le 1. \tag{18}$$

where $\gamma$ is a high frequency boost factor. The images reconstructed using zero-filled partial k-space for different $\gamma$ values are shown in panels (a1)-(a4) of Fig. 10. Effect of incidental phase artifacts is seen to increase with boost factor. In addition to this, Gibbs ringing is prominently seen as periodic repetitions of the circular edge in all the images. Panels (b)-(d) show images reconstructed using homodyne, extended homodyne and POCS filters respectively. It is seen that with increasing value of $\gamma$, the reduction of incidental phase artifacts are more obvious in both the homodyne methods. Also, the rate of change of intensity with high frequency boosting provides an indication of the filter's effectiveness. The uniformity of intensity levels for the extended homodyne filter, as seen from the intensity versus $\gamma$ plots in Fig. 11 confirms this.

Fig. 10

Fig. 11

A quantitative evaluation of the three phase correction methods is performed by calculating the error in the reconstructed image using

$$error = \frac{\| S_{original} - S_{partial} \|^2}{\| S_{original} \|^2} \quad (19)$$

where $S_{original}$ and $S_{partial}$ are the images reconstructed from the full k-space and partial k-space respectively. Fig. 12 shows the dependence of error on the percentage of acquired k-space for reconstruction using zero filling, POCS filter, homodyne filter, and the proposed extended homodyne filter. It is seen that the error decreases with increase in the percentage of acquired k-space, and minimum for the extended homodyne filter.

Fig. 12

Similar performance can be seen for 3D reconstruction applied to PCMRA in Fig. 13. It is seen that the error decreases with increase in the fraction of acquired k-space, and minimum for the modified homodyne filter.

Fig. 13

## Acknowledgement

The authors wish to thank the Department of Science and Technology (DST SR/S3/EECE/0107/2012) of India for scholarship and operating funds. We also thankfully acknowledge Dr. Jaladhar Neelavalli, Assistant Professor and Uday Bhaskar Krishnamurthy, Research Assistant, Department of Radiology, Wayne State University, Michigan for providing the in-vivo MR data. We would like to extend our thanks to Dr. Sairam Geethanath, director Rashmi reddy, Biomedical research centre, Dayanand sagar institution for providing PCMRA data.

## References

[1] Z. P. Liang, F. E. Boda, R. T. Constable, E. M. Haacke, P. C. Lauterbur, and M.R. Smith, "Constrained reconstruction methods in MR imaging," Reviews Magn. Reson. Med.,vol. 4, pp. 67-185, 1992.

[2] M. L. Wood and R. M. Henkelman, "Truncation artefacts in magnetic resonance imaging," Magnetic Resonance in Medicine, vol. 2, no. 6, pp. 517-526, Feb. 1985.

[3] T. B. Smith, and K. S. Nayak, "MRI artifacts and correction strategies," Imaging Med. Review, vol. 2, no. 4, pp. 445-457, 2010.

[4] O. Dietrich, M. F. Reiser, and S. O. Schoenberg, "Artifacts in 3-Tesla MRI: Physical background and reduction strategies," Eur J Radiol, vol. 65, pp. 29-35, 2007.

[5] H. Schomberg, "Off-resonance correction of MR images," IEEE Trans.Med.Imag.,vol. 18, no. 6, pp. 481-495, June 1999.

[6] Ping Hou, Khader M. Hasan, Clark W. Sitton, Jerry S. Wolinsky, and Ponnada A. Narayana, "Phase-Sensitive T1 Inversion Recovery Imaging: A Time-Efficient Interleaved Technique for Improved Tissue Contrast in Neuroimaging," AJNR Am J Neuroradiol, vol.26, pp.1432–1438, 2005

[7] A. Carlsson, "Susceptibility effects in MRI and H MRS," Doctoral Thesis, Dept. Radiation Physics, University of Gothenburg, SE, Sweden, 2009.

[8] D.C.Noll, D.G.Nishimura, A.Macovski, "Homodyne detection in magnetic resonance imaging," *IEEE Trans Med Imaging*, vol.10, pp.154–163, 1991.

[9] J. Pauly, "Partial k-space reconstruction,"

http//users.fmrib.ox.ac.uk/~karla/reading_group/lecture_notes/Recon_Pauly_read.pdf

[10] J. Chen, L. Zhang, Y. Zhu, and J. Luo, "MRI reconstruction from 2D partial k-space using POCS algorithm," in *Int. Conf. on Bioinformatics and Biomedical Engineering*, Beijing, 2009, pp. 1-4.

[11] N. J. PeIc, M. A. Bemstein, A. Shimakawa, G. H. Glovem, "Encoding strategies for three-direction phase-contrast MR imaging of flow," *J Magn Reson Imaging*, vol.1, pp. 405-413, 1991.

[12] M. Blaimer, F. Breuer, M. Mueller, R. M. Heidemann, M. A. Griswold, P. M. Jakob, "SMASH, SENSE, PILS, GRAPPA How to Choose the Optimal Method," *Magn Reson Imaging*, vol.15, pp. 223-236, 2004.

## Figure Captions

**Fig. 1(A): Fourier reconstruction applied to partial k-space of SWI. (a) Sketch of full k-space, corresponding Fourier reconstructed magnitude and phase images, (b) Sketch of 2D partial k-space with *n* fractional lines in phase encode direction and *m* frequency points in the frequency encode direction, corresponding Fourier reconstructed magnitude and phase images. Insets show the Regions of Interest (ROIs) exhibiting signal losses.**

**Fig. 1(B): Fourier reconstruction applied to partial k-space of GREI. (a) Sketch of full k-space, corresponding Fourier reconstructed magnitude and phase images, (b) Sketch of 2D partial k-space with *n* fractional lines in phase encode direction and *m* frequency points in the frequency encode direction, corresponding Fourier reconstructed magnitude and phase images. Insets show the Region of Interest (ROI) with IPA.**

**Fig. 2 : Simulation using numerical Phantom. (a) Magnitude image, (b) Low frequency phase image, (c) Phase image with combination of both low and high frequency components.**

**Fig. 3: Fourier reconstruction applied to partial k-space of simulated data. (a) Image reconstructed from the k-space of magnitude image in Fig. 2(a) after 2D truncation, (b) Image reconstructed from the k-space of magnitude image in Fig. 2(a) in combination with phase image of 2(b) after 2D truncation, (c) Image reconstructed from the k-space of magnitude image in Fig. 2(a) in combination with phase image of 2(c) after 2D truncation.**

**Fig. 4: Block schematic of the proposed method.**

**Fig. 5: Application of phase correction methods to GREI data. (a1) Image reconstructed from 2D partial k-space, (a2) ROI within the red bounding box, (b1) Image obtained by application of POCS method to the 2D partial k-space, (b2) ROI within the red bounding box, (c1) Image obtained by application of Conventional homodyne method to the 2D partial k-space, (c2) ROI within the red bounding box, (d1) Image obtained by application of extended homodyne method to the 2D partial k-space, (d2) ROI within the red bounding box. Yellow contours show the areas exhibiting signal losses.**

**Fig. 6: Comparison of phase correction methods applied to SWI data. (a) Image reconstructed from 2D partial k-space, (a1)-(a3) ROIs within the colored bounding boxes, (b) Image obtained by application of POCS method to the 2D partial k-space, (b1)-(b3) ROIs within the colored bounding boxes, (c) Image obtained by application of Conventional homodyne method to the 2D partial k-space, (c1)-(c3) ROIs within the colored bounding boxes, (d) Image obtained by application of extended homodyne method to the 2D partial k-space, (d1)-(d3) ROIs within the colored bounding boxes.**

**Fig. 7. Illustration of 3D partial k-space**

**Fig. 8A : Comparison of various Phase correction methods applied to MIP images reconstructed using four-point method from velocity encoded 3D partial k-space partitions (dataset#1). a1) zero-filled reconstruction, (b1) reconstruction using 3D POCS, (c1) reconstruction using extended 3D homodyne method, (a2)-(c2) ROIs within the red bounding box.**

**Fig. 8B : Comparison of various Phase correction methods applied to MIP images reconstructed using four-point method from velocity encoded 3D partial k-space partitions (dataset#2). a1) zero-filled reconstruction, (b1) reconstruction using 3D POCS, (c1) reconstruction using extended 3D homodyne method, (a2)-(c2) ROIs within the red bounding box.**

**Fig. 9A : Image reconstruction using 2D truncated coil k-spaces with undersampling along phase encode direction. (a1)-(a4): Individual channel images reconstructed using zero-filled k-space, (b1)-(b4): Individual channel images reconstructed using GRAPPA , (c1)-(c4), (d1)-(d4), (e1)-(e4) : Individual channel images reconstructed using POCS, homodyne, extended homodyne applied to GRAPPA filled k-spaces in (b), (a)-(e) Combined coil images using each method. Yellow arrows show the areas exhibiting artifacts and signal losses.**

**Fig. 9B : Same as Fig. 9A each panel corresponds to the regions around the yellow arrow shown in Fig. 9A. The panel descriptions are same as that in Fig.9A**

**Fig. 10 : Comparison of various phase correction methods applied to numerically simulated data. The magnitude image is constructed using a circular object with uniform intensity. The spatial low frequency phase is simulated using Eq.(1) and the high frequency component is simulated using a set of concentric discs with uniform phase within each segment and abrupt transition of phase across segments. The amount of incidental phase variation is numerically simulated using Eq. (18) for different values of the high**

**frequency boost factor $\gamma$. The images shown correspond to row-wise increasing values of $\gamma$ indicating higher levels of incidental phase variation. Column-wise panels illustrate the different reconstruction methods. (a1)-(a4) images reconstructed using zero-filled partial k-space, (b1)-(b4) homodyne, (c1)-(c4) extended homodyne, and (d1)-(d4) POCS method.**

**Fig. 11 : Plot of Intensities versus high frequency boost factor ($\gamma$).**

**Fig. 12: Reconstruction error measured with reference to the image reconstructed from full k-space versus percentage of acquired k-space**

**Fig. 13 : Reconstruction error measured with reference to the maximum intensity projections of image volumes obtained using 3D reconstruction of full k-space partitions versus fraction of acquired k-space.**

# Figures

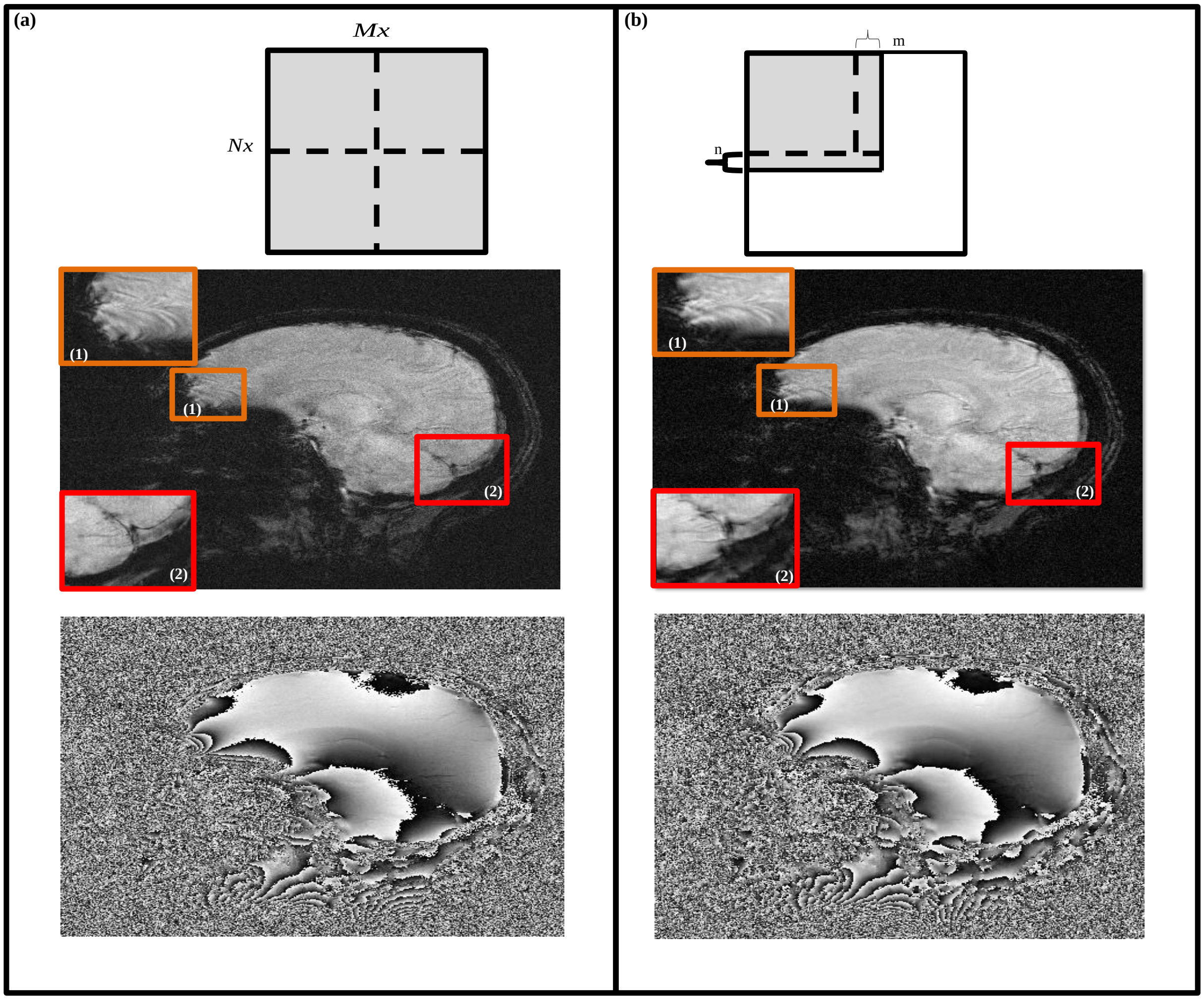

**Fig. 1(A): Fourier reconstruction applied to partial k-space of SWI.**

**(a) Sketch of full k-space, corresponding Fourier reconstructed magnitude and phase images, (b) Sketch of 2D partial k-space with *n* fractional lines in phase encode direction and *m* frequency points in the frequency encode direction, corresponding Fourier reconstructed magnitude and phase images. Insets show the Regions of Interest (ROIs) exhibiting signal losses.**

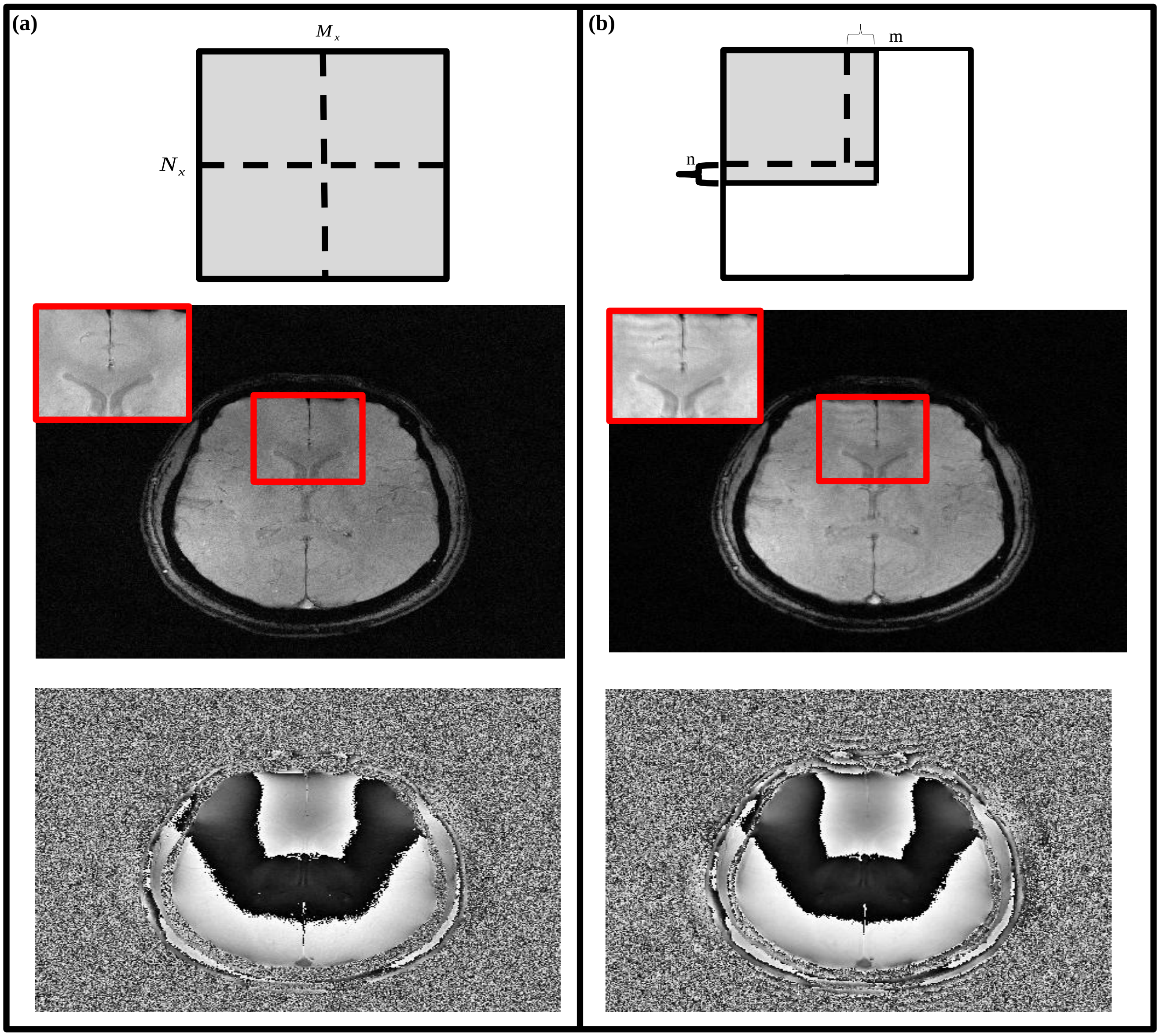

**Fig. 1(B): Fourier reconstruction applied to partial k-space of GREI. (a) Sketch of full k-space, corresponding Fourier reconstructed magnitude and phase images, (b) Sketch of 2D partial k-space with *n* fractional lines in phase encode direction and *m* frequency points in the frequency encode direction, corresponding Fourier reconstructed magnitude and phase images. Insets show the Region of Interest (ROI) with IPA.**

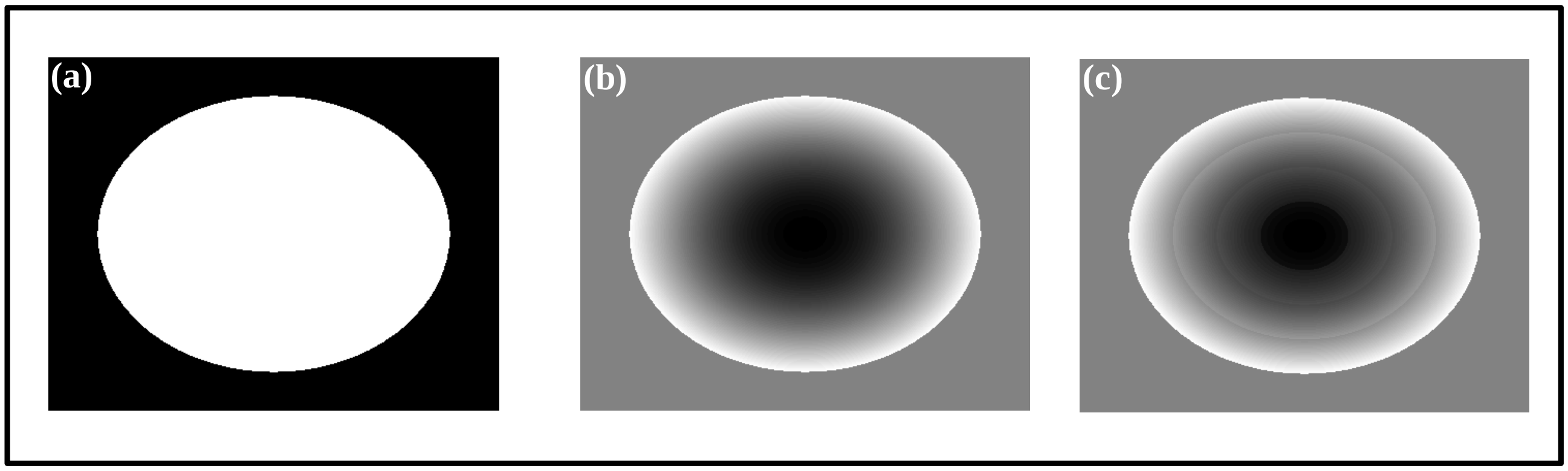

**Fig. 2 : Simulation using numerical Phantom. (a) Magnitude image, (b) Low frequency phase image, (c) Phase image with combination of both low and high frequency components.**

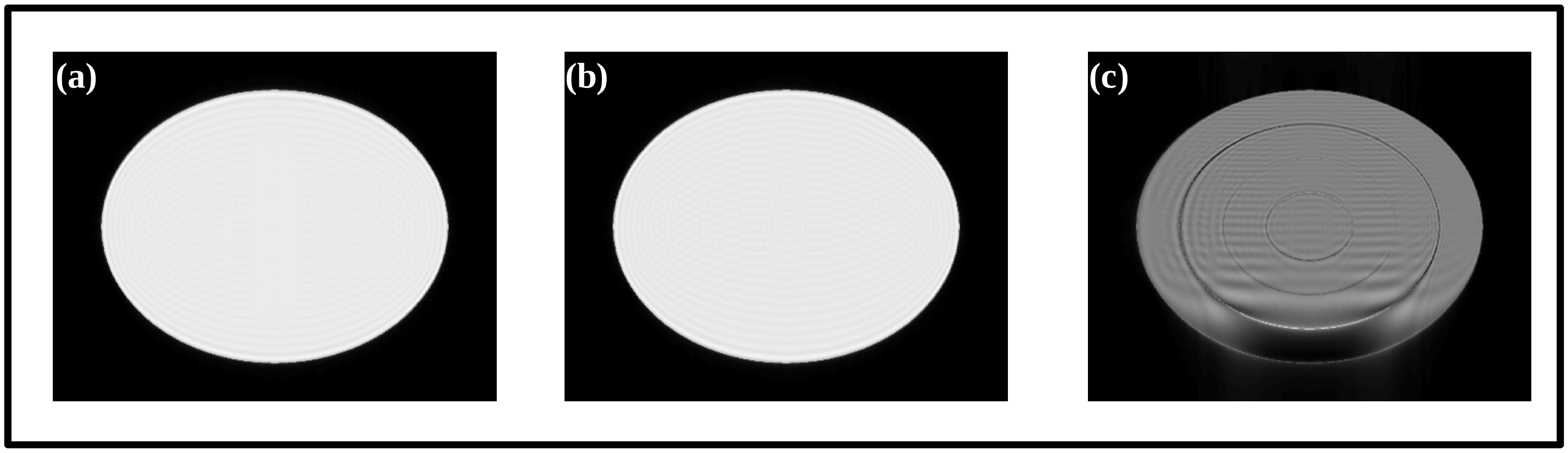

**Fig. 3: Fourier reconstruction applied to partial k-space of simulated data. (a) Image reconstructed from the k-space of magnitude image in Fig. 2(a) after 2D truncation, (b) Image reconstructed from the k-space of magnitude image in Fig. 2(a) in combination with phase image of 2(b) after 2D truncation, (c) Image reconstructed from the k-space of magnitude image in Fig. 2(a) in combination with phase image of 2(c) after 2D truncation.**

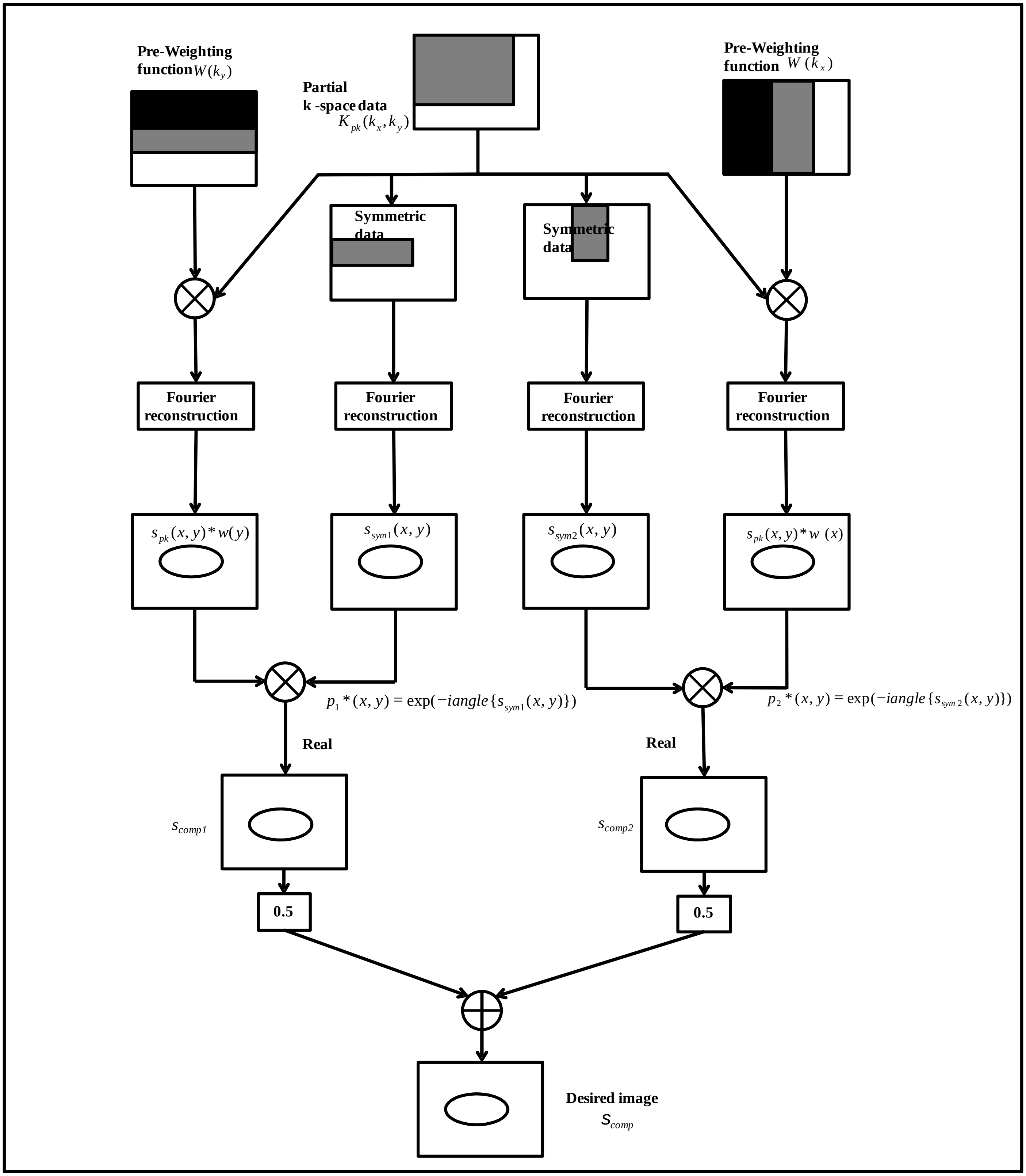

**Fig. 4: Block schematic of the proposed method.**

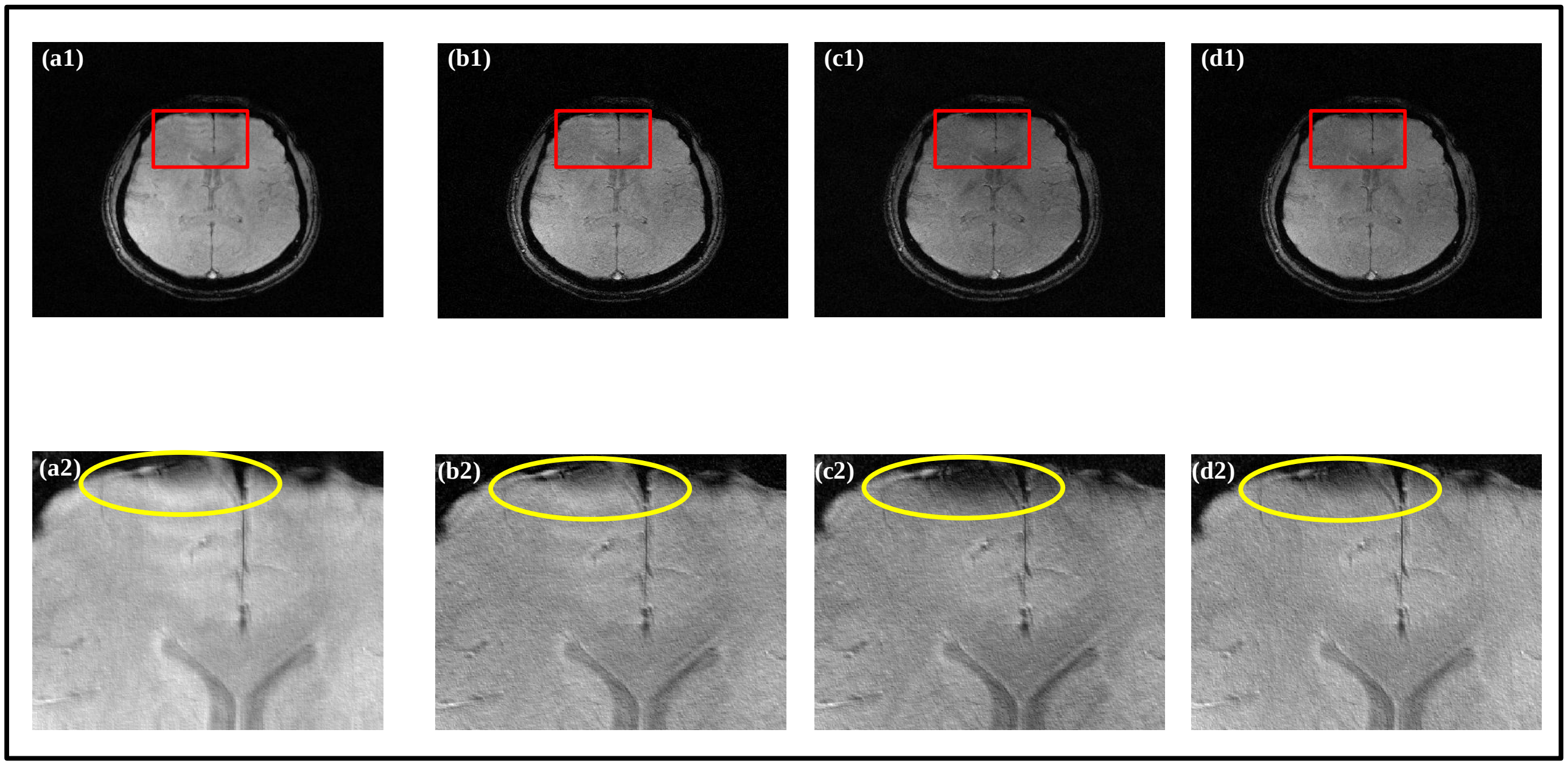

**Fig. 5: Application of phase correction methods to GREI data. (a1) Image reconstructed from 2D partial k-space, (a2) ROI within the red bounding box, (b1) Image obtained by application of POCS method to the 2D partial k-space, (b2) ROI within the red bounding box, (c1) Image obtained by application of Conventional homodyne method to the 2D partial k-space, (c2) ROI within the red bounding box, (d1) Image obtained by application of extended homodyne method to the 2D partial k-space, (d2) ROI within the red bounding box. Yellow contours show the areas exhibiting signal losses.**

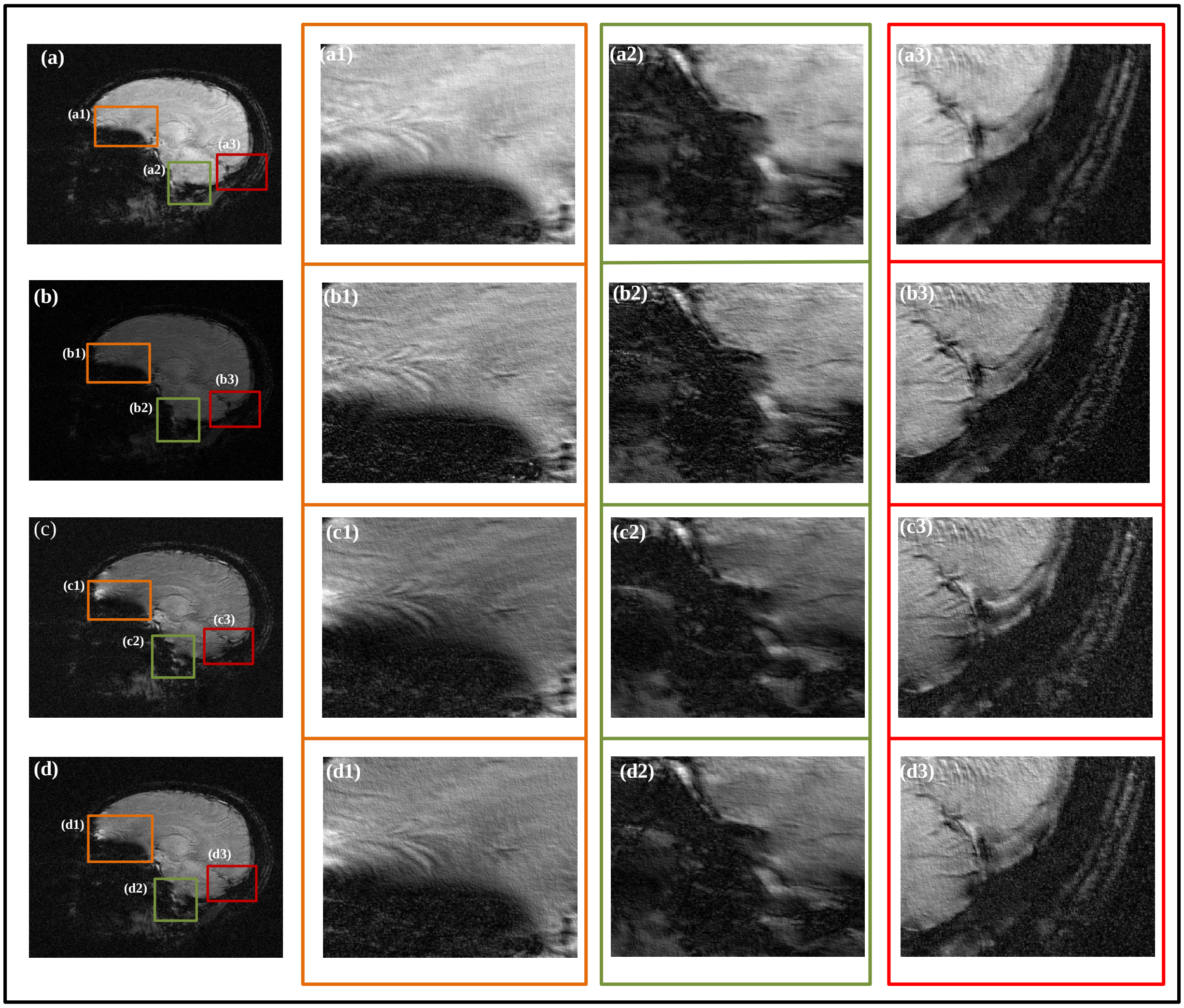

**Fig. 6: Comparison of phase correction methods applied to flow compensated SWI data. (a) Image reconstructed from 2D partial k-space, (a1)-(a3) ROIs within the colored bounding boxes, (b) Image obtained by application of POCS method to the 2D partial k-space, (b1)-(b3) ROIs within the colored bounding boxes, (c) Image obtained by application of Conventional homodyne method to the 2D partial k-space, (c1)-(c3) ROIs within the colored bounding boxes, (d) Image obtained by application of extended homodyne method to the 2D partial k-space, (d1)-(d3) ROIs within the colored bounding boxes.**

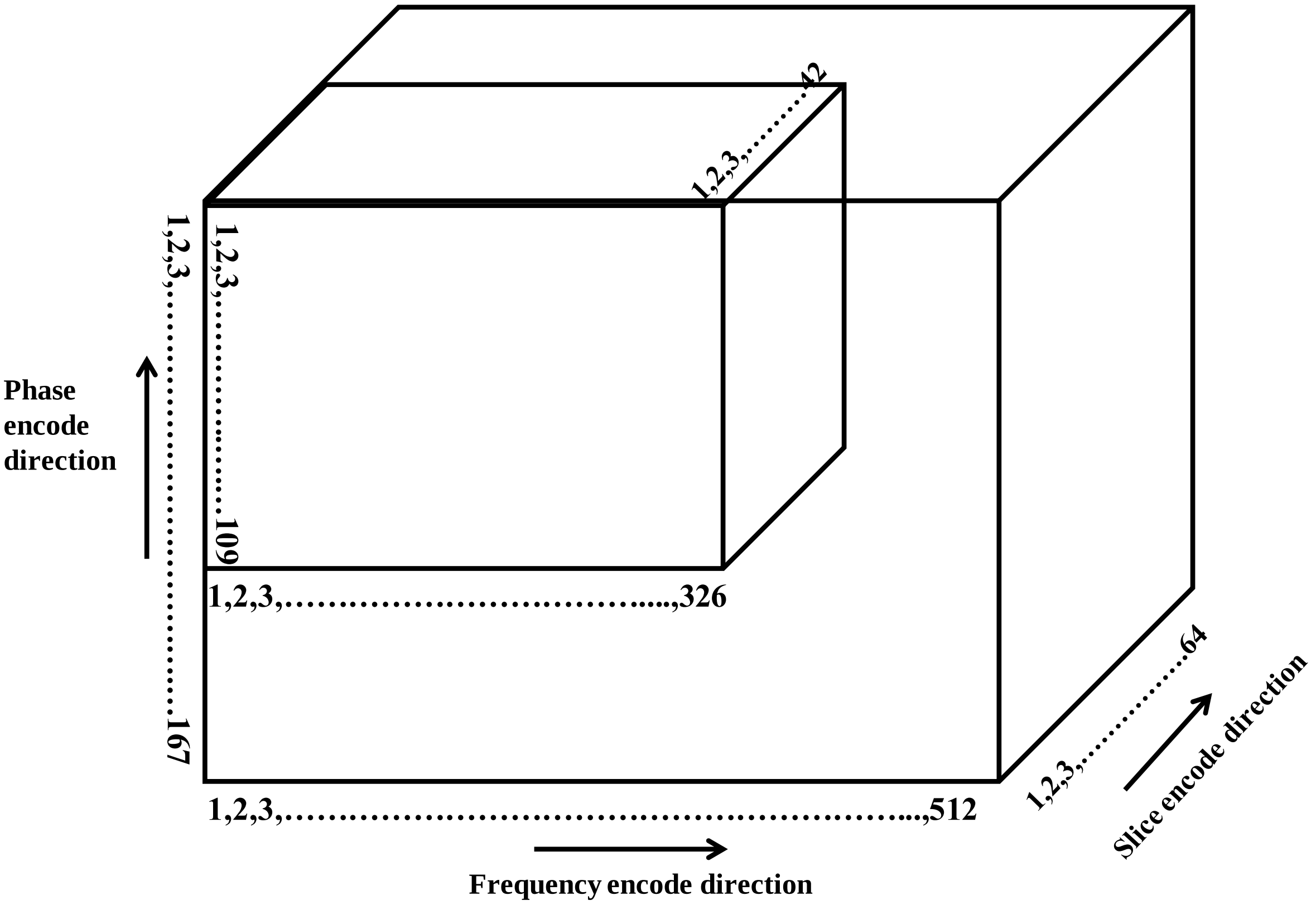

**Fig. 7. Illustration of 3D partial k-space**

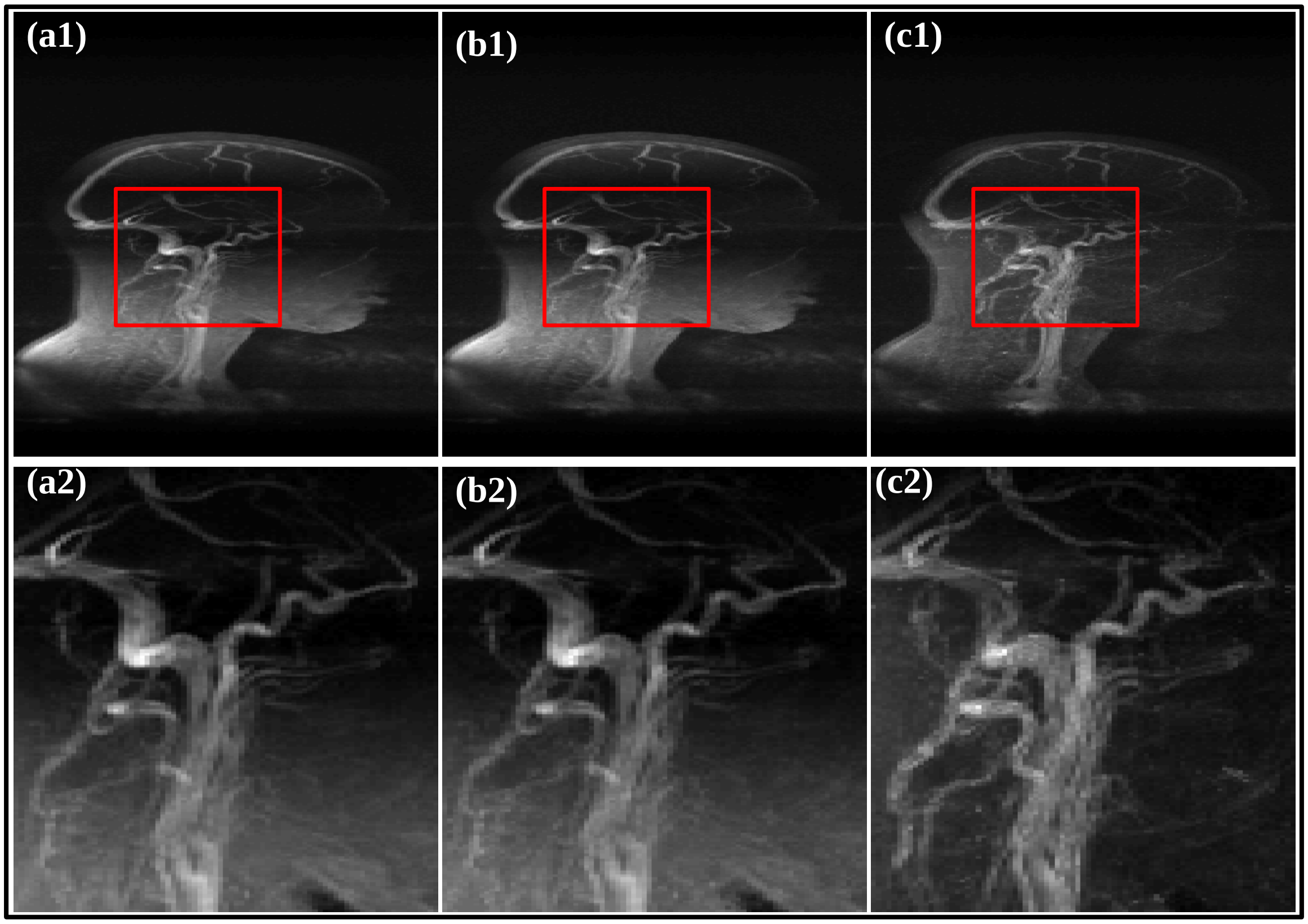

**Fig. 8A : Comparison of various Phase correction methods applied to MIP images reconstructed using four-point method from velocity encoded 3D partial k-space partitions (dataset#1). a1) zero-filled reconstruction, (b1) reconstruction using 3D POCS, (c1) reconstruction using extended 3D homodyne method, (a2)-(c2) ROIs within the red bounding box.**

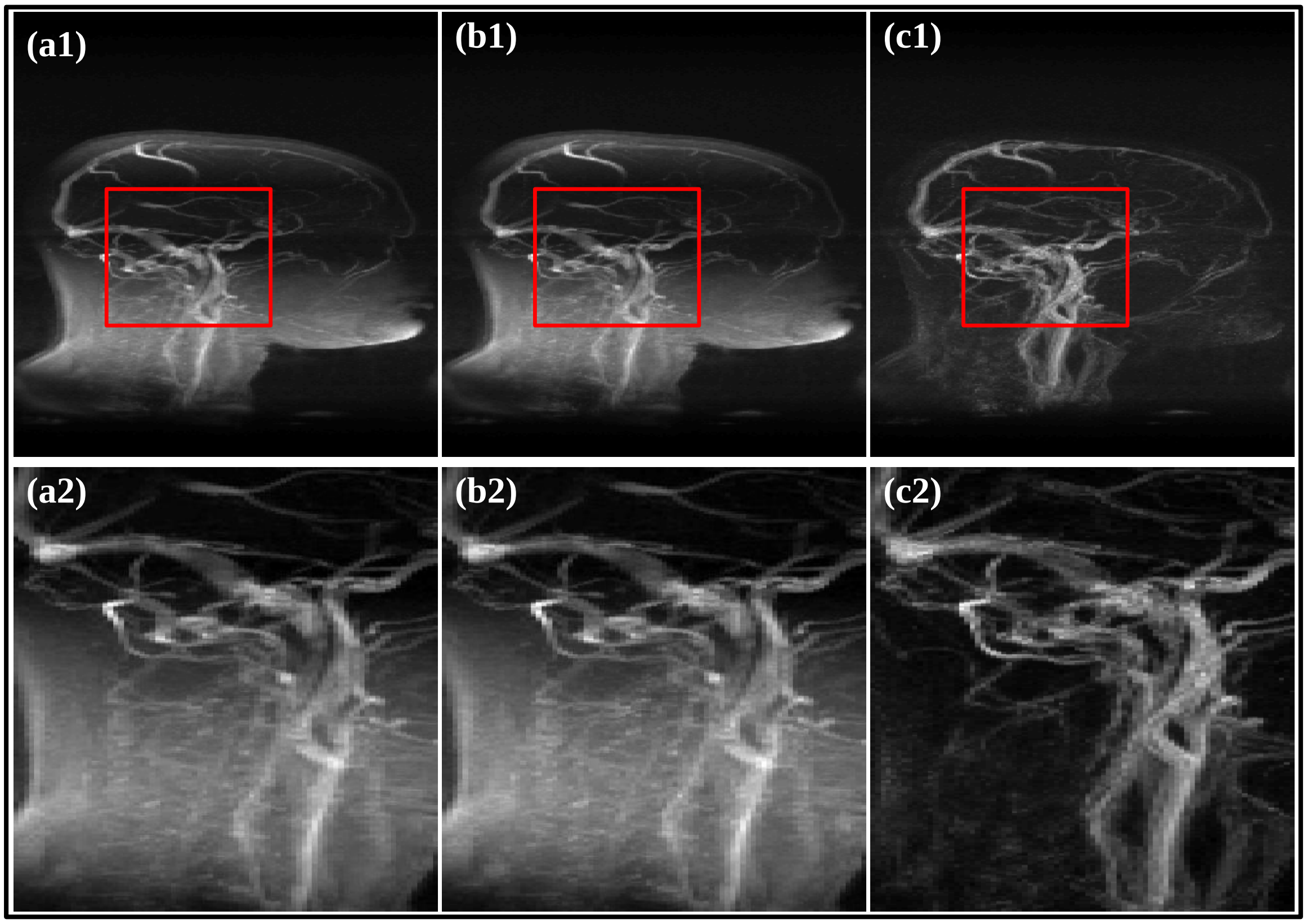

**Fig. 8B : Comparison of various Phase correction methods applied to MIP images reconstructed using four-point method from velocity encoded 3D partial k-space partitions (dataset#2). a1) zero-filled reconstruction, (b1) reconstruction using 3D POCS, (c1) reconstruction using extended 3D homodyne method, (a2)-(c2) ROIs within the red bounding box.**

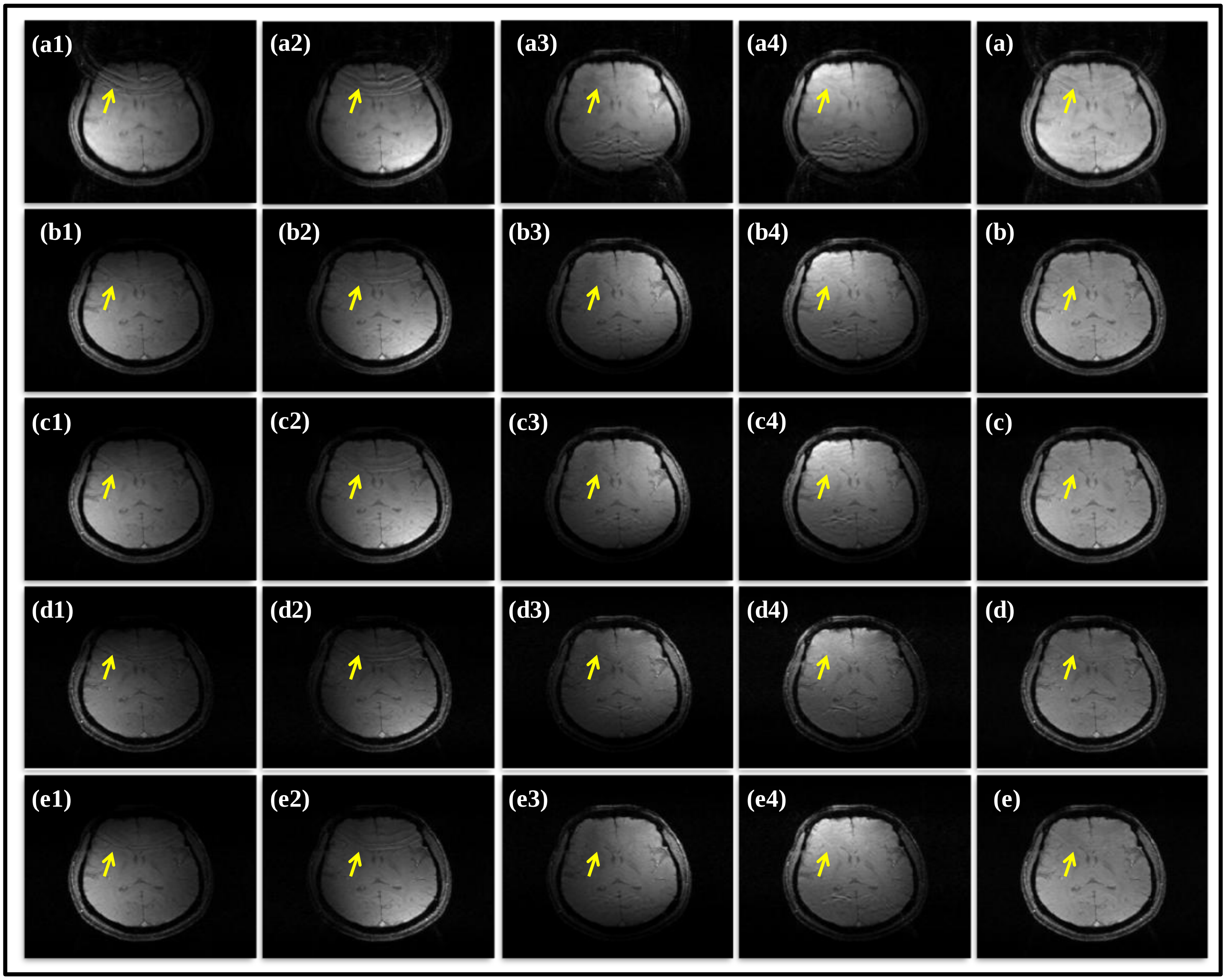

**Fig. 9A : Image reconstruction using 2D truncated coil k-spaces with undersampling along phase encode direction. (a1)-(a4): Individual channel images reconstructed using zero-filled k-space, (b1)-(b4): Individual channel images reconstructed using GRAPPA , (c1)-(c4), (d1)-(d4), (e1)-(e4) : Individual channel images reconstructed using POCS, homodyne, extended homodyne applied to GRAPPA filled k-spaces in (b), (a)-(e) Combined coil images using each method. Yellow arrows show the areas exhibiting artifacts and signal losses.**

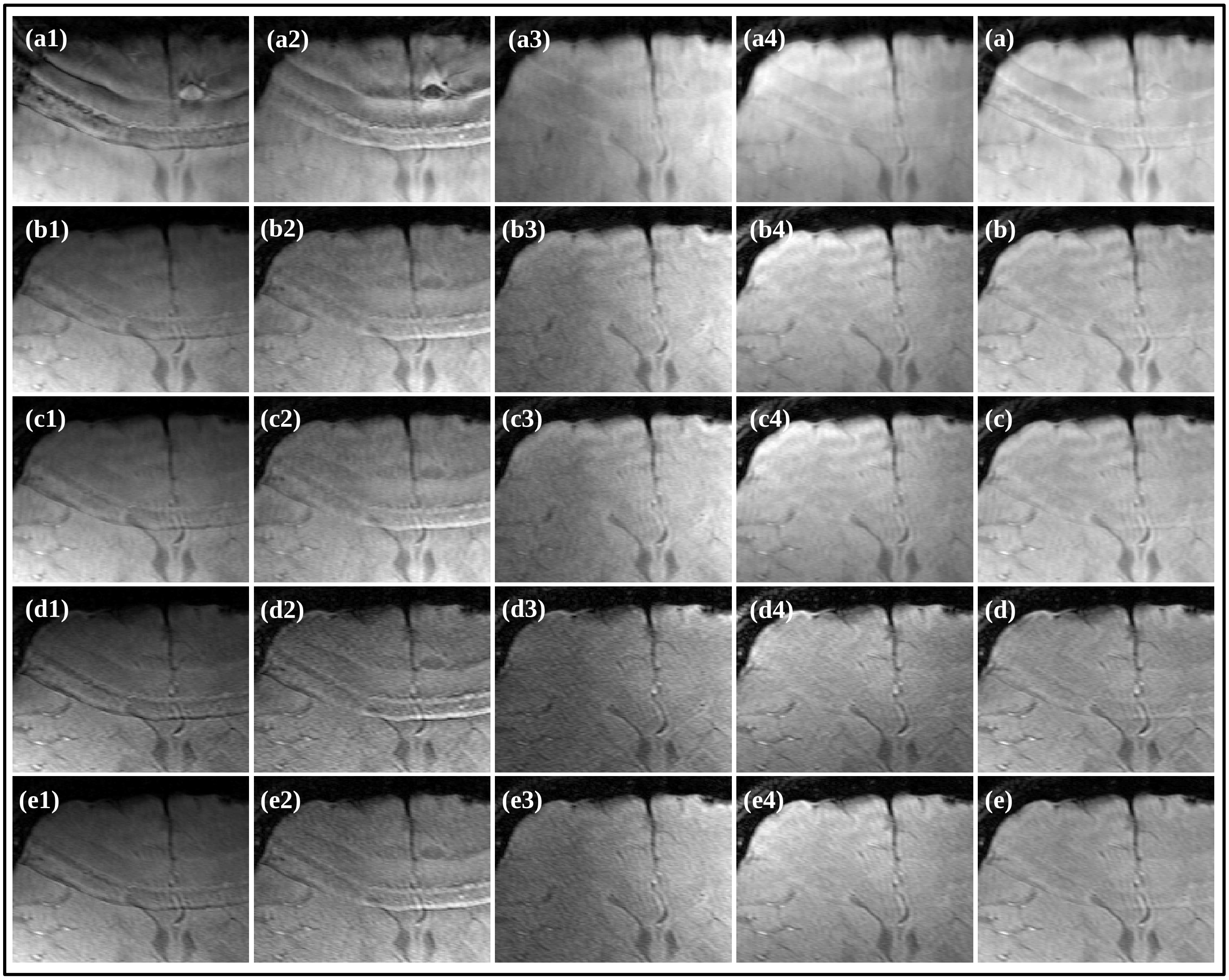

**Fig. 9B : Same as Fig. 9A each panel corresponds to the regions around the yellow arrow shown in Fig. 9A. The panel descriptions are same as that in Fig.9A**

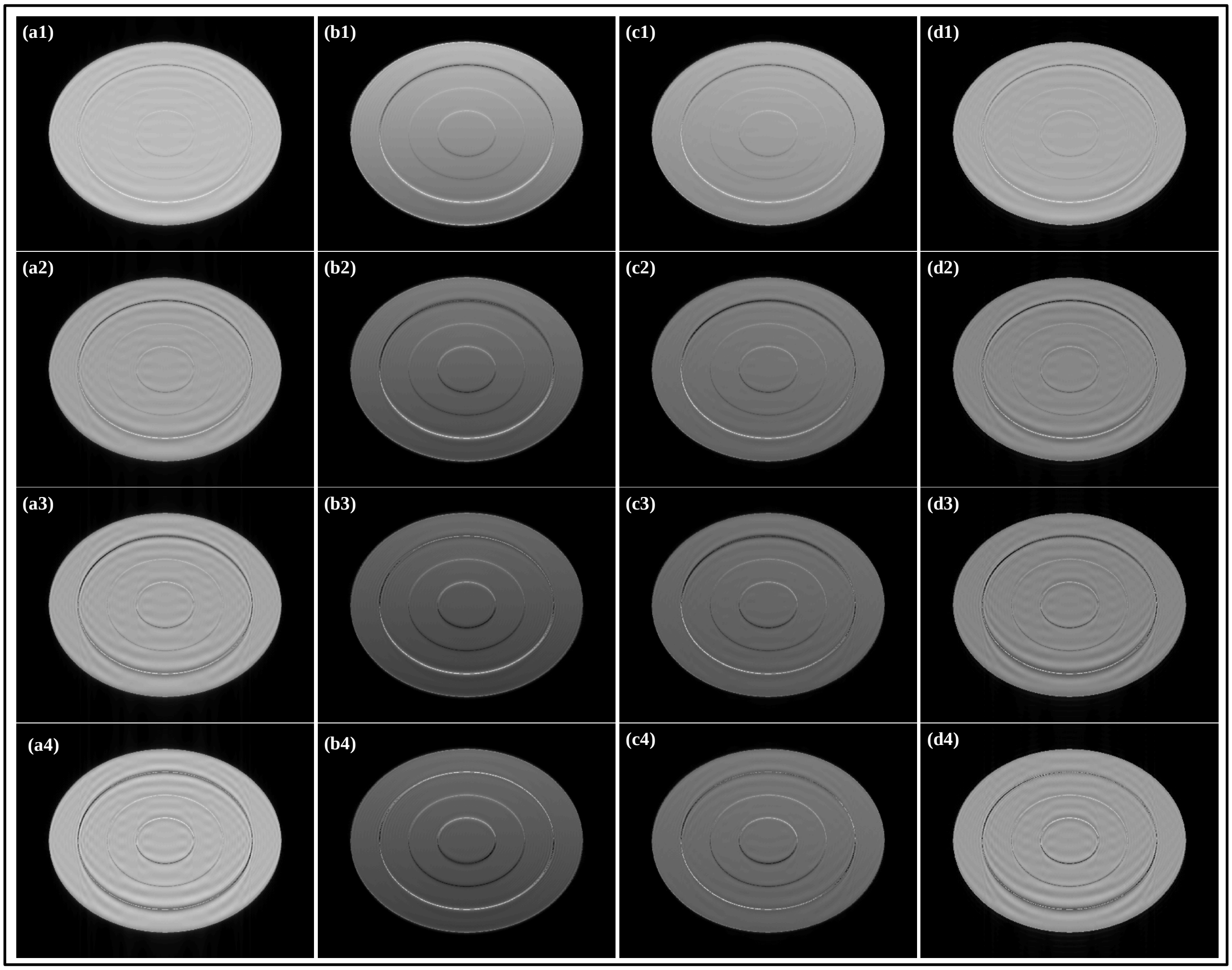

**Fig. 10 : Comparison of various phase correction methods applied to numerically simulated data. The magnitude image is constructed using a circular object with uniform intensity. The spatial low frequency phase is simulated using Eq.(1) and the high frequency component is simulated using a set of concentric discs with uniform phase within each segment and abrupt transition of phase across segments. The amount of incidental phase variation is numerically simulated using Eq. (18) for different values of the high frequency boost factor $\gamma$. The images shown correspond to row-wise increasing values of $\gamma$ indicating higher levels of incidental phase variation. Column-wise panels illustrate the different reconstruction methods. (a1)-(a4) images reconstructed using zero-filled partial k-space, (b1)-(b4) homodyne, (c1)-(c4) extended homodyne, and (d1)-(d4) POCS method.**

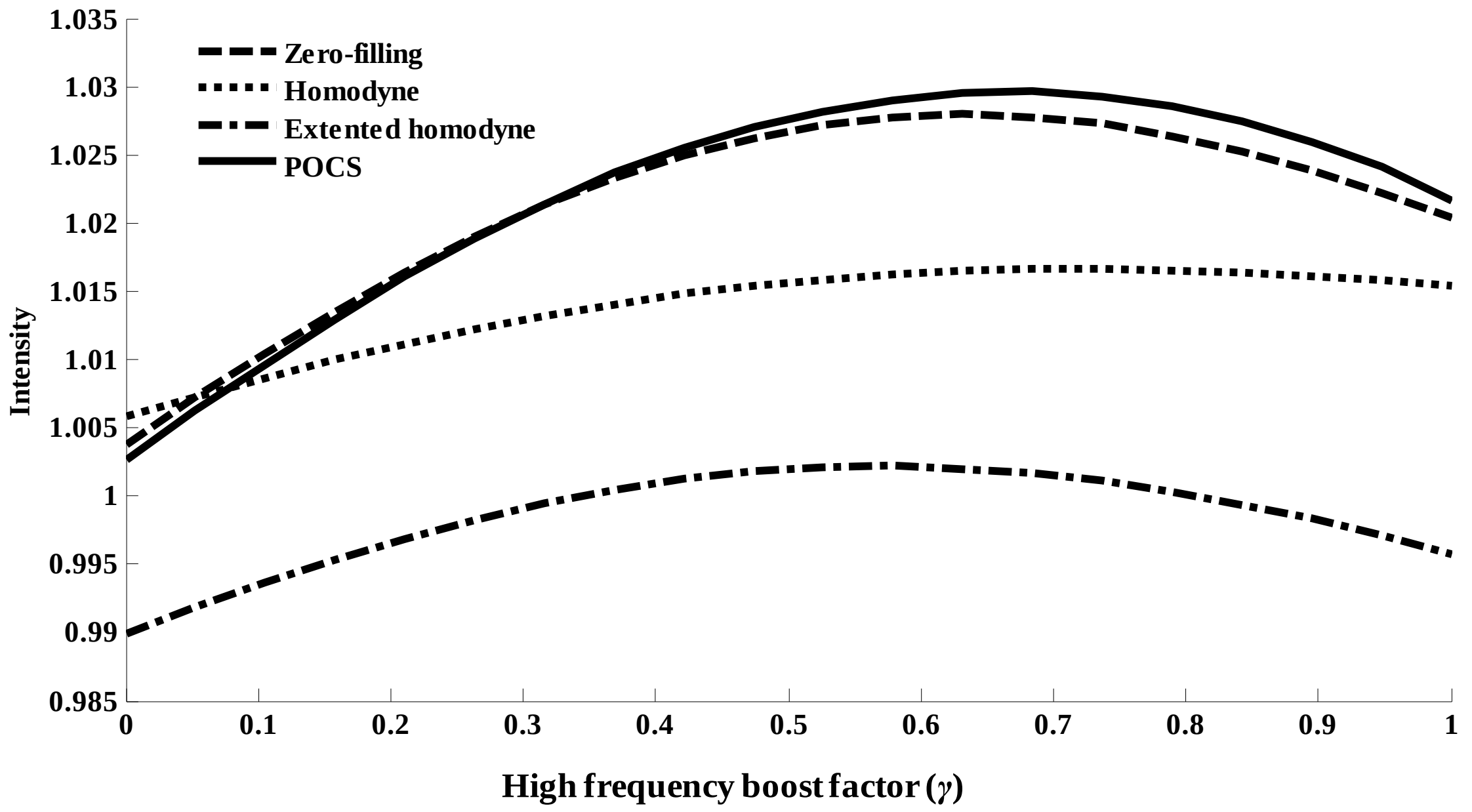

**Fig. 11 : Plot of Intensities versus high frequency boost factor (*γ*)**

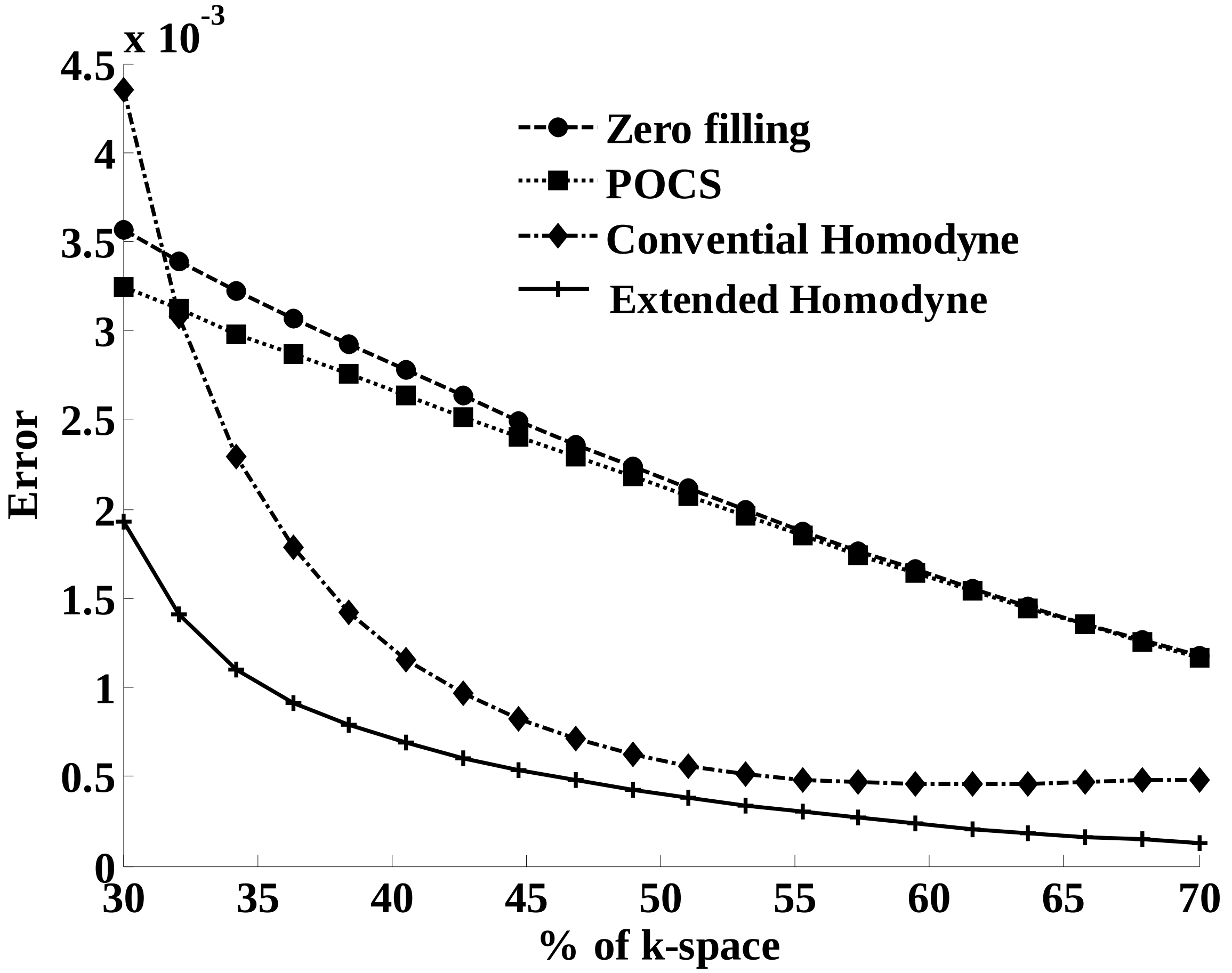

**Fig. 12: Reconstruction error measured with reference to the image reconstructed from full k-space versus percentage of acquired k-space**

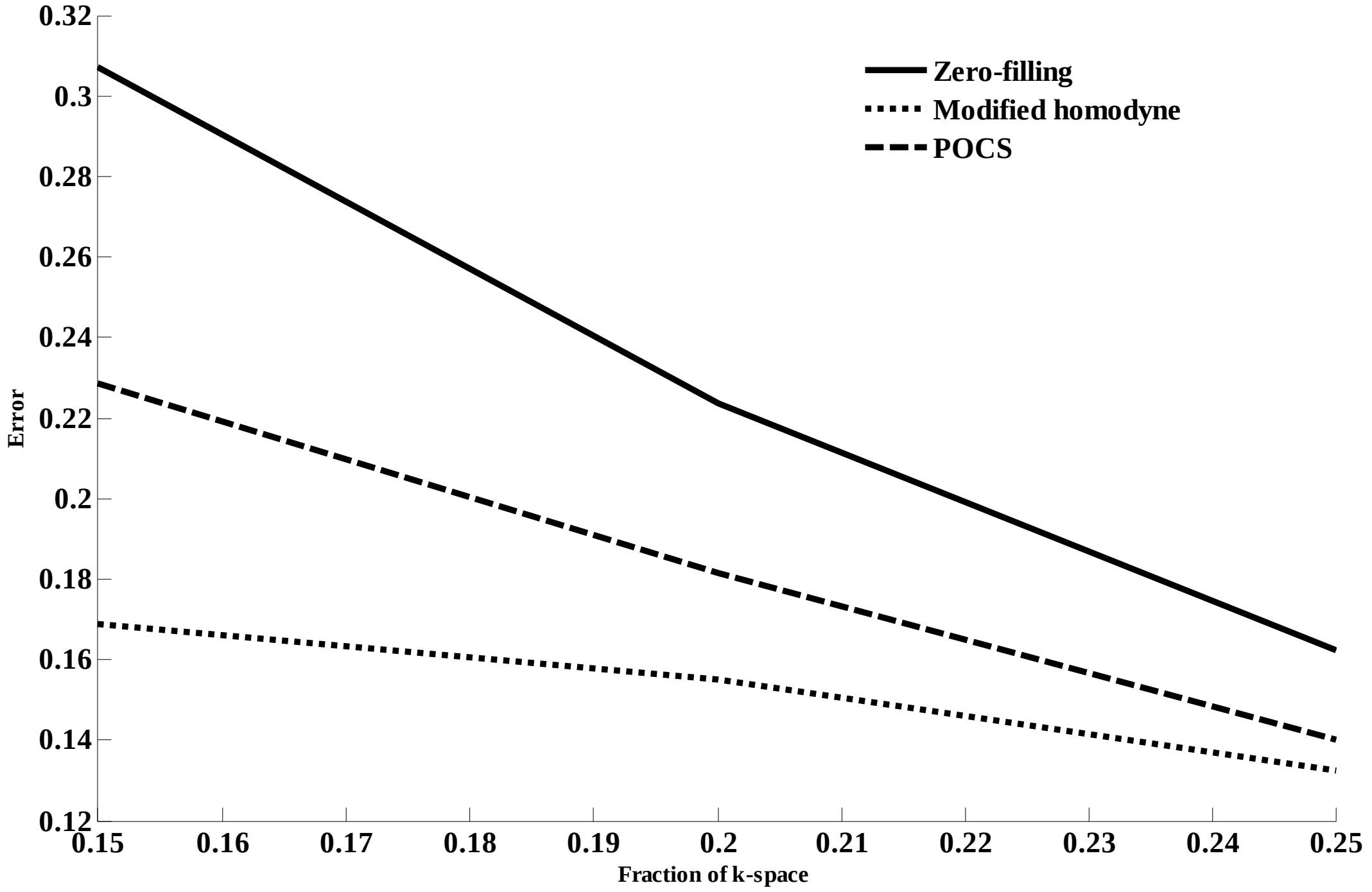

**Fig. 13 : Reconstruction error measured with reference to the maximum intensity projections of image volumes obtained using 3D reconstruction of full k-space partitions versus fraction of acquired k-space.**